\documentclass[10pt,twocolumn,letterpaper]{article}

\usepackage{cvpr} 
\usepackage{times}
\usepackage{epsfig}
\usepackage{graphicx}
\usepackage{amsmath}
\usepackage{amssymb}
\usepackage{float}
\usepackage{multirow}
\usepackage [latin1]{inputenc}

\begin{document}

\title{On the Use of Synthetic Data for Threshold Calibration in Face Recognition: Performance and Security Implications for Border Control Systems}

\author{Arto Apila\\
Faculty of Information Technology and Electrical Engineering, University of Oulu, Finland\\
{\tt\small arto.apila@oulu.fi}}

\maketitle
\thispagestyle{empty}


\begin{abstract}
The recently deployed Entry/Exit System (EES) introduces large-scale biometric verification into European border control, requiring face recognition systems to operate at extremely low false match rates (FMR). While regulatory frameworks define performance targets at the EES Central System level, they do not specify how verification thresholds should be calibrated in practice at the Member State level. In operational settings, obtaining representative real-world data for calibration is often constrained by legal, logistical, and privacy limitations.

In this work, we investigate the use of synthetic face data for threshold calibration in document-to-live verification scenarios relevant to border control systems. We analyze the alignment of genuine and impostor score distributions between synthetic and real datasets and evaluate the transferability of calibrated thresholds across domains, with a focus on low-FMR operating points.

Our results show that synthetic data can approximate calibration behavior in controlled settings, but fails to reliably generalize to unconstrained conditions due to mismatches in score distribution tails. These discrepancies lead to significant degradation in recognition performance and increased vulnerability to morph-based attacks. We further demonstrate that calibration outcomes are highly dataset-dependent, even across synthetic datasets.

Overall, our findings highlight that while synthetic data is useful for system development and preliminary calibration, our results indicate that reliable threshold selection in high-security deployments typically requires validation and adjustment using representative real-world data.

\end{abstract}

\section{Introduction}

The Entry/Exit System (EES), mandated by Regulation (EU) 2017/2226 \cite{ref_ees_regulation_2017}, introduces large-scale biometric verification in European border control. It captures facial images and fingerprints of third-country nationals and performs identity verification by matching stored references against live acquisitions.

While the EES Central System and shared Biometric Matching Service (sBMS) are operated by eu-LISA, Member States are responsible for implementing local verification functions within their border control systems \cite{ref_ees_regulation_2017,ref_sbms_2025_launch}.

Frontex guidelines require local document-to-live face verification prior to enrollment to prevent identity fraud using forged documents. This aims to prevent impostors from being enrolled using forged documents \cite{ref_frontex_ees_tech_guide_2021,ref_ees_ec_overview}.

Regulations define performance targets (e.g., False Negative Matching Rate (FNMR) $<$ 1\% at False Matching Rate (FMR) = 0.05\%) for the eu-LISA's (sBMS) used in EES, but do not specify how verification thresholds should be calibrated (i.e., setting the accept/reject decision threshold for face recognition)  at the Member State level \cite{ref_ees_biometrics_2019}. Earlier guidelines for automated border control recommend FNMR $<$ 5\% and FMR = 0.1\% \cite{ref_frontex_abc_tech_guidelines_2015}, which are no longer aligned with modern system performance.

While synthetic face datasets have been shown to support model training and benchmarking \cite{ref_colbois_2021,ref_borsukiewicz_2025_synthetic}, their suitability for threshold calibration in operational settings remains largely unexplored. In particular, it is unclear whether synthetic data can accurately reproduce the score distributions required for reliable threshold selection under strict low-FMR constraints, as required in systems such as EES. Recent European policy work recognizes the potential of synthetic biometric datasets but recommends that representative operational data remain essential for biometric accuracy evaluation \cite{eulisa2026biometric}.

We investigate, for the first time, whether thresholds derived from synthetic datasets generalize to real-world document-to-live verification scenarios.

Our contributions are:
\begin{itemize}
    \item Analysis of genuine and impostor score distribution alignment between synthetic and real datasets.
    \item Evaluation of cross-dataset threshold generalization under low-FMR constraints.
    \item Assessment of system robustness using morph-based attack scenarios.
\end{itemize}

Importantly, we show that even small discrepancies, particularly under domain shift in score distributions, can lead to significant operational risks at low-FMR operating points. Experiments are conducted using a lightweight EdgeFace model to reflect realistic deployment constraints. To our knowledge, this is the first systematic analysis of threshold calibration transfer using synthetic data under low-FMR constraints relevant to operational biometric systems.

\section{Related Work}

This section reviews prior work on face recognition datasets, synthetic data for training and evaluation, and challenges in threshold calibration under low-FMR constraints.

\subsection{Face Recognition and Large-Scale Datasets}

Deep face recognition systems have achieved state-of-the-
art performance through the use of large-scale datasets and discriminative learning objectives. Modern face recognition systems rely on margin-based losses such as ArcFace \cite{arcface} and large-scale labeled datasets. Large-scale datasets such as LFW \cite{ref_lfw}, VGGFace2 \cite{ref_vggface2}, and WebFace260M \cite{ref_webface260m} have been central to training and benchmarking modern systems. Many of these datasets have been restricted due to privacy and legal concerns, limiting reproducibility \cite{ref_borsukiewicz_2025_synthetic}. 

However, existing work has not systematically investigated whether synthetic data can support reliable threshold calibration under strict low-FMR constraints in operational settings.

Synthetic face data has emerged as a scalable alternative to real datasets. Prior work demonstrates its effectiveness for training, including DigiFace-1M \cite{ref_digiface_2023}, SynFace \cite{ref_synface_2021}, and hybrid approaches combining synthetic and real data \cite{ref_granoviter_2023,ref_variface_2024}. Large-scale evaluations further show that well-designed synthetic datasets can achieve competitive recognition performance while preserving key properties such as intra-class variability and identity separability \cite{ref_borsukiewicz_2025_synthetic}.

Synthetic data has also been explored for benchmarking, with studies showing that it can reproduce relative model rankings and error trends \cite{ref_colbois_2021}. Initiatives such as SDFR \cite{ref_sdfr_2024} further highlight its potential as a scalable evaluation resource. However, discrepancies in absolute score distributions between synthetic and real data remain a limitation for applications requiring precise calibration.

\subsection{Threshold Calibration and Low-FMR Operation}

In operational face recognition systems, performance is defined at specific operating points corresponding to target false match rates (FMR). Threshold calibration therefore depends on accurate estimation of genuine and impostor score distributions \cite{ref_jain_biometrics_book}. However, thresholds are highly sensitive to dataset characteristics and domain shifts, and often fail to generalize across datasets \cite{ref_klare_ijcb2012,ref_nguyen_domain_shift,liu2022oneface}. Despite its importance, threshold calibration remains underexplored, particularly in the context of synthetic data.

Face recognition performance is governed by the distributions of genuine and impostor similarity scores, which define operating thresholds and error rates. While synthetic datasets can approximate these distributions \cite{ref_colbois_2021}, variations in image quality, pose, and illumination can lead to deviations that significantly impact calibration, particularly at low-FMR operating points \cite{Wu2023BMVCBalanced}.

\section{Methodology}

We design an experimental framework to evaluate the effectiveness of synthetic data for threshold calibration under operational constraints. The framework focuses on (i) score distribution alignment across synthetic and real datasets, (ii) cross-dataset threshold transfer at low-FMR operating points, and (iii) robustness under morph-based attacks.

\subsection{Datasets}

In operational face recognition systems, threshold calibration at low-FMR operating points depends critically on accurate modeling of impostor score distribution tails, yet it remains unclear whether synthetic data can provide sufficient fidelity for this purpose under realistic domain shifts.
We use both synthetic and real-world datasets to evaluate threshold calibration and generalization. A morph image dataset is used to analyze the security implications of decision-threshold calibration and to evaluate how transferable these results are to morph attack detection.

\textbf{Synthetic datasets.} FLUXSyn-ID \cite{ref_fluxsynid_2025} is a synthetic dataset containing 14,889 identities with paired document-style and live images, enabling document-to-live verification. Images are generated using a diffusion-based pipeline with identity-controlled conditioning and variations in pose, illumination, and expression. Multiple live images per identity are generated using complementary methods that introduce controlled and unconstrained variations while preserving identity consistency.

Compared to prior synthetic datasets, FLUXSyn-ID has been shown to achieve improved alignment with real-world data distributions in embedding space and similarity score distributions, as well as higher inter-class diversity \cite{ref_fluxsynid_2025}. These properties make it particularly suitable for evaluating biometric systems under operational constraints. Unlike unconstrained synthetic datasets, FLUXSyn-ID explicitly models the document-to-live domain shift, which is critical for threshold calibration in real-world verification systems. However, prior analysis has shown that certain attributes, such as age and fine-grained facial features, may not be consistently represented, and similarity-based filtering can introduce shifts in demographic distributions \cite{ref_fluxsynid_2025}.

\begin{figure}[h]
\centering
\includegraphics[width=0.6\linewidth]{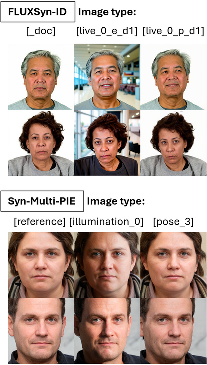}
\caption{Examples of synthetic images by dataset and image type.}
\label{fig:synth}
\end{figure}

In this study, the FLUXSyn-ID image types \texttt{f\_doc} (synthetic document-style reference image), \texttt{live\_0\_e\_d1} (identity-preserving live image with controlled variations), and \texttt{live\_0\_p\_d1} (live image with more unconstrained, natural variations) are used for calibration and evaluation. Genuine (mated) pairs are formed by comparing each image of an identity with all other images of the same identity. Fig.~\ref{fig:synth} illustrates example images from the FLUXSyn-ID dataset.

Syn-Multi-PIE \cite{ref_colbois_2021} is a synthetic dataset generated using a StyleGAN2-based framework to replicate the characteristics of the Multi-PIE dataset. It provides controlled variations in pose, illumination, and expression by manipulating latent directions in the generative model. The dataset was specifically designed for benchmarking face recognition systems under controlled conditions and has been shown to produce comparable system rankings and error trends to real datasets, although with reduced intra-class variability.

The generation process of Syn-Multi-PIE relies on latent space editing to preserve identity while introducing controlled variations, enabling systematic evaluation of robustness to nuisance factors such as pose and lighting \cite{ref_colbois_2021}.

In this study, the Syn-Multi-PIE image types \texttt{reference} (neutral expression, frontal pose, and frontal illumination), \texttt{illumination\_0} (same identity under controlled illumination changes), and \texttt{pose\_3} (same identity under non-frontal pose variations) are used for calibration and evaluation. Genuine (mated) pairs are formed by comparing each image of an identity with all other images of the same identity. Fig.~\ref{fig:synth} illustrates example images from the Syn-Multi-PIE dataset.

\textbf{Real-world datasets.} Color FERET \cite{ref_feret_nist, ref_feret_program} provides high-quality images captured under controlled conditions, resulting in low intra-class variability and well-separated score distributions. Such conditions are consistent with ICAO-compliant image acquisition standards used in travel documents and guided capture processes in systems such as EES, making FERET a suitable proxy for controlled enrollment and verification scenarios. CFPW \cite{cfp-paper} introduces unconstrained conditions with large facial expression and illumination variations, enabling evaluation under realistic domain shifts. Table~\ref{tab:dataset_stats} summarizes the characteristics of the calibration and evaluation datasets.

\begin{figure}[h]
\centering
\includegraphics[width=0.5\linewidth]{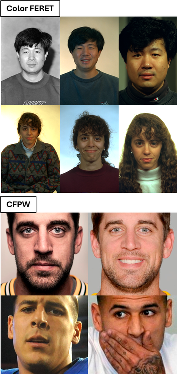}
\caption{Examples of Color FERET  and CFPW frontal images.}
\label{fig:real}
\end{figure}

In this study, Color FERET image types \texttt{\_fa} and \texttt{\_fb} (frontal images), and CFPW image type \texttt{\_frontal}, are used for evaluation. Genuine (mated) pairs are formed by comparing each image of an identity with all other images of the same identity. Color FERET provides 2--10 frontal images per identity, while CFPW provides 20. Fig.~\ref{fig:real} illustrates example images from the Color FERET and CFPW datasets.

\begin{figure}[h]
\centering
\includegraphics[width=0.6\linewidth]{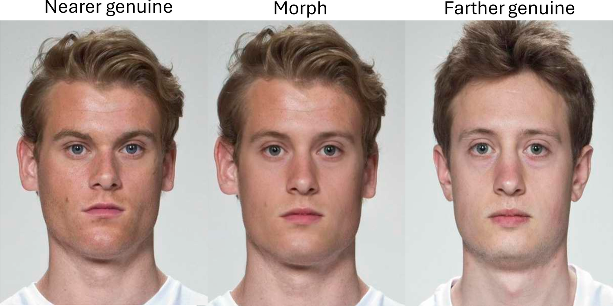}
\caption{Example AMSL morph images.}
\label{fig:MORPH}
\end{figure}

\textbf{Morph images.} To evaluate robustness under attack conditions, we use morph images (AMSL dataset) generated from the Face Research Lab London Set \cite{ref_debruine_2017} using a landmark-based blending approach \cite{ref_neubert_2018}. The dataset consists of 2,175 morphed images generated from pairs of genuine images belonging to 102 identities. Each morph image is compared against both contributing identities. Fig.~\ref{fig:MORPH} illustrates example images from the AMSL morph dataset.

\textbf{Pair generation.}
For each dataset, genuine (mated) pairs are formed by comparing images of the same identity. Impostor (non-mated) pairs are generated by comparing images across different identities. Genuine pairs are formed from images of the same identity, while impostor pairs are randomly sampled across identities to obtain representative distributions.

\begin{table}[H]
\begin{center}
\setlength{\tabcolsep}{4pt}
\begin{tabular}{lccccc}
\hline
Dataset   & IDs  & Gen.  & Imp. & Gen. $\bar{x}$ & Imp. $\bar{x}$\\
\hline
FLUXSyn.  & 14889 & 44653 & 100M & 0.832         &  0.107\\
Syn-Multi.& 10000 & 30000 & 100M & 0.807         &  0.199\\
Synth.Comb.  & 24889 & 74653 & 200M & 0.822         &  0.099\\
CFPW      & 500   & 21735 & 49M  & 0.692         & 0.015\\
Col. FERET  & 994 &  3652 & 3.7M & 0.876         & 0.015\\
\hline
\end{tabular}
\end{center}
\caption{\label{tab:dataset_stats}Calibration and evaluation dataset statistics. Synth.Comb. = combined dataset of synthetic datasets, IDs = identities, Gen. = genuine pairs, Imp. = impostor pairs.}
\end{table}

\begin{figure}[H]
\centering
\includegraphics[width=0.60\linewidth]{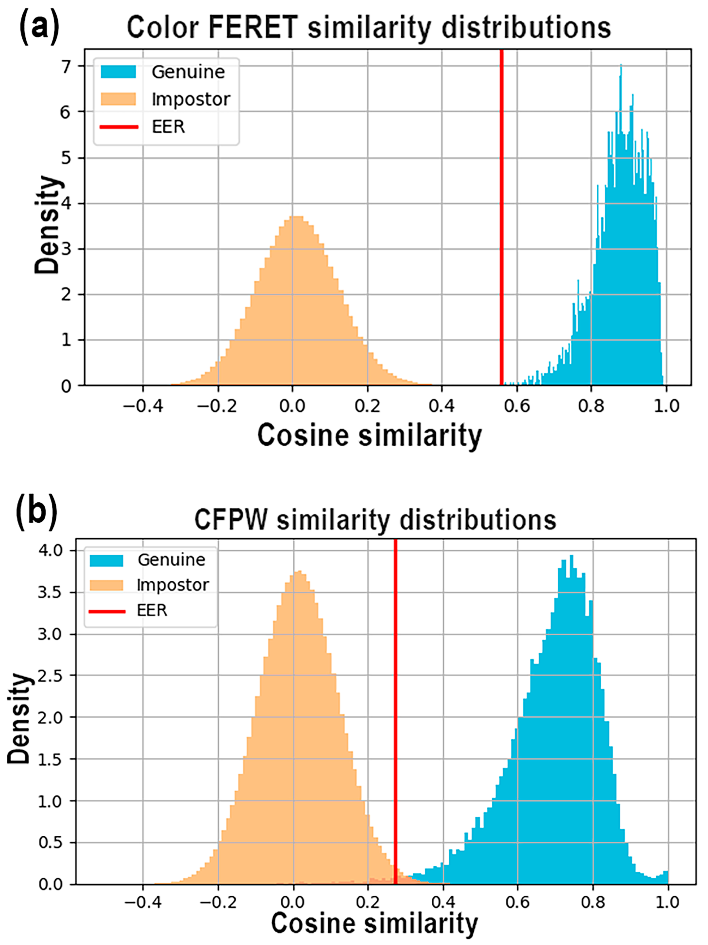}
\caption{Distance distributions on (a) Color FERET frontal subset and on (b) CFPW subset.}
\label{fig:real_distributions}
\end{figure}

\subsection{Face Recognition Model}

All experiments are conducted using the EdgeFace model \cite{ref_edgeface_2024}, a lightweight deep face recognition architecture designed for efficient deployment on edge devices while maintaining competitive recognition performance. Performance comparison is presented in Table \ref{tab:lfw_comp}. The developers of EdgeFace trained the model on selected subsets of the WebFace260M dataset.

We use the \textbf{EdgeFace-S} variant with scaling parameter $\gamma = 0.5$ (EdgeFace\_S\_gamma\_05), which provides a favorable trade-off between computational efficiency and recognition accuracy. This configuration follows the recommendations of \cite{ref_edgeface_2024} for resource-constrained environments.

\begin{table}[h]
\begin{center}
\setlength{\tabcolsep}{4pt}
\begin{tabular}{lccc}
\hline
Model & Params & LFW Acc. (\%)  \\
\hline
DeepID3 & $\sim$100M+ & 99.53  \\
FaceNet & $\sim$140M & 99.63  \\
ArcFace (ResNet100) & $\sim$65M & 99.83  \\
MagFace (ResNet100) & $\sim$65M & 99.83\\
\hline
MobileFaceNet & $\sim$1.0M & 99.28 \\
EdgeFace-XS ($\gamma = 0.6$)  & 1.77M & 99.73  \\
\textbf{EdgeFace-S ($\gamma = 0.5$)} & 3.65M & 99.80  \\
\hline
\end{tabular}
\end{center}
\caption{Comparison of lightweight and high-end face recognition models on LFW. Reported results are taken from original publications: DeepID3 \cite{Deedip}, FaceNet \cite{facenet}, ArcFace \cite{arcface}, MagFace \cite{magface}, MobileFaceNet \cite{mobilefacenet} and EdgeFace \cite{ref_edgeface_2024}.}
\label{tab:lfw_comp}
\end{table}

The implementation strictly follows the pipeline described in the original work. Facial images are aligned using standard preprocessing procedures, and embeddings are extracted from the final feature layer of the network. The model is used as provided by the authors, without additional fine-tuning, to ensure reproducibility and avoid dataset-specific bias.

EdgeFace is particularly suitable for this study as it reflects realistic deployment constraints in border control systems, where computational efficiency, latency, and scalability are critical factors. At the same time, it provides sufficiently strong recognition performance to enable meaningful analysis of threshold calibration and security behavior. While higher-capacity models may achieve lower absolute error rates, the use of EdgeFace allows us to focus on calibration behavior under realistic deployment constraints, where model efficiency and stability are as important as peak accuracy.

\subsection{Calibration and evaluation Protocols}

Threshold calibration is performed independently on each calibration dataset using impostor score distributions. Thresholds are estimated from impostor score distributions as quantiles corresponding to target FMR levels. We consider operating points ranging from $0.1\%$ to $0.001\%$, reflecting typical requirements in border control applications.

Evaluation is conducted on separate datasets to assess the generalization of calibrated thresholds (i.e., no image overlap is permitted between datasets used for calibration). For each evaluation dataset, we compute Cosine similarity scores for both genuine and impostor pairs. We report False Non-Match Rate (FNMR) and observed False Match Rate (FMR). This allows us to assess how well thresholds calibrated on one dataset transfer to another. In addition to real-world evaluation, synthetic datasets are also used in controlled cross-dataset experiments to analyze calibration transfer under known distribution shifts. Synthetic datasets are used exclusively for controlled transfer experiments to analyze distributional effects and are not intended to represent operational evaluation conditions.

\subsection{Morph Attack Evaluation}

For each morph image, similarity scores are computed against both contributing identities. We evaluate whether the morph is accepted as either or both identities under a given threshold.

We report two standard metrics:

\begin{itemize}
    \item \textbf{NMPR (Nearer Match Passing Rate):} proportion of morphs accepted as the more similar identity.
    \item \textbf{FMPR (Farther Match Passing Rate):} proportion of morphs accepted as the less similar identity.
\end{itemize}

These metrics quantify the vulnerability of the system to morph-based attacks under different calibration strategies and operating points.

\noindent\textbf{Scope and operational realism.}
While this study is motivated by border control systems such as EES, the experimental setup focuses specifically on the problem of threshold calibration under controlled variations in data distributions, rather than modeling the full operational pipeline. In practice, deployed systems include additional factors such as sensor variability, capture quality constraints, demographic balancing, and multi-stage decision logic, which are not explicitly represented in our experiments. Instead, our goal is to isolate the effect of dataset-dependent score distributions on calibration behavior, particularly in the low-FMR regime where decision thresholds are determined by the tails of impostor distributions. The selected datasets therefore represent controlled and unconstrained extremes, allowing us to study calibration transfer under varying degrees of domain shift. While this abstraction does not fully capture all operational conditions, it enables a focused analysis of a critical component of deployment that remains largely underexplored.

\section{Results}
We first analyze calibration thresholds (Table \ref{tab:thresholds}), followed by cross-dataset evaluation on real data (Tables \ref{tab:fnmr} and \ref{tab:fnmr_CFPW}), and finally bidirectional transfer between synthetic datasets (Tables \ref{tab:fnmr_Syn} and \ref{tab:fnmr_Flux}). The results reveal significant inconsistencies in threshold transferability, particularly under low-FMR constraints. Synthetic-to-synthetic experiments are included to isolate distributional effects under controlled conditions and should be interpreted as analytical experiments rather than operational evaluations.

\subsection{Calibration}

The observed differences in calibrated thresholds across synthetic datasets can be explained by variations in score distributions, which are closely linked to intrinsic dataset properties such as intra-class variability and identity separability. Prior work has shown that synthetic datasets differ significantly in these characteristics, depending on their generation process and design objectives \cite{ref_borsukiewicz_2025_synthetic}. 

\begin{figure}[H]
\centering
\includegraphics[width=0.62\linewidth]{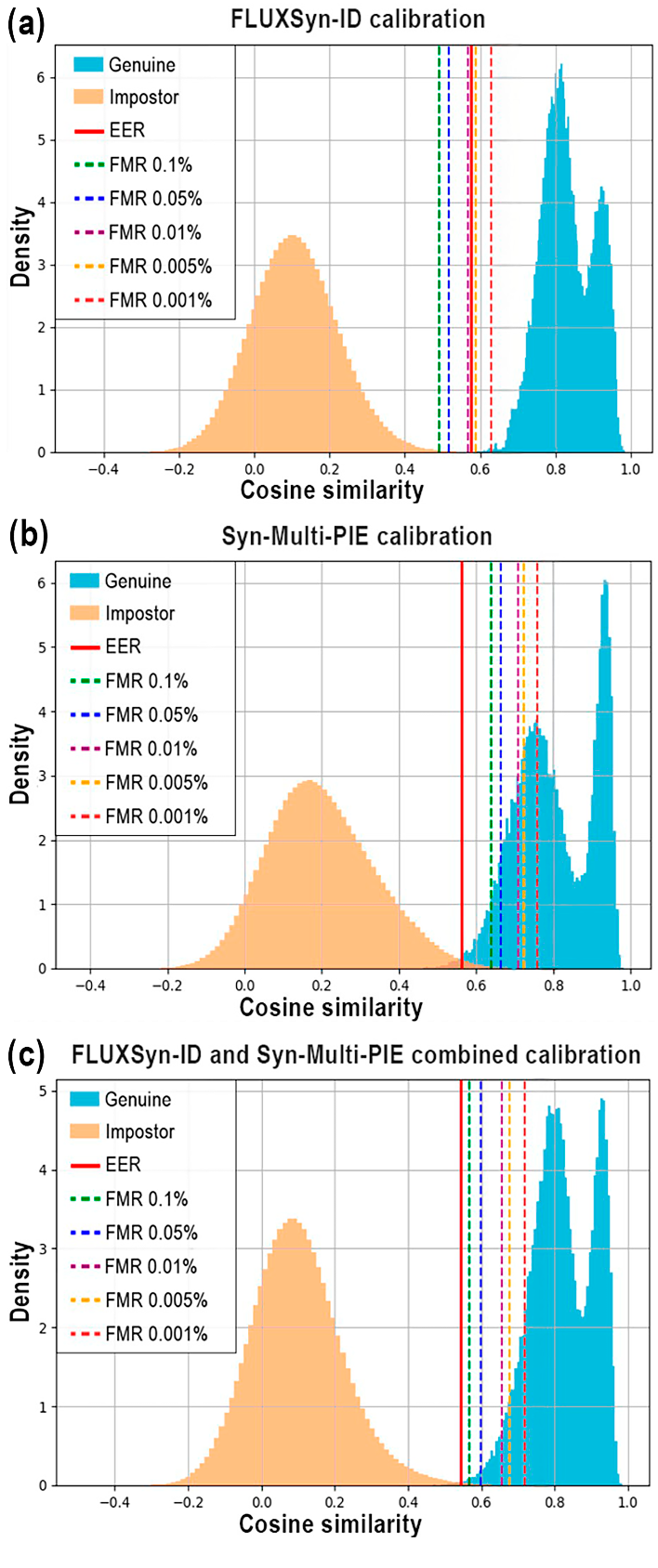}
\caption{Distance distributions and calculated thresholds per FMR of different calibration datasets. a) FLUXSyn-ID, b) Syn-Multi-PIE, c) FLUXSyn-ID and Syn-Multi-PIE combined.}
\label{fig:distributionsFLUX}
\end{figure}

Fig.~\ref{fig:distributionsFLUX} illustrates the underlying score distributions used for calibration, highlighting differences in impostor tail behavior that directly influence threshold selection. Table \ref{tab:thresholds} shows the exact thresholds (cosine similarity) derived from each dataset.

\begin{table}[H]
\begin{center}
\setlength{\tabcolsep}{4pt}
\begin{tabular}{lccccc}
\hline
Dataset & 0.10\% & 0.05\% & 0.01\% & 0.005\% & 0.001\% \\
\hline
FLUXSyn. & 0.492 & 0.518 & 0.570 & 0.590 & 0.630   \\
Syn-Multi. & 0.640 & 0.663 & 0.708 & 0.708 & 0.725 \\
Combined & 0.567 & 0.597 & 0.656 & 0.678  & 0.719   \\
\hline
\end{tabular}
\end{center}
\caption{\label{tab:thresholds}Decision thresholds (cosine similarity) at selected FMR levels.}
\end{table}

Datasets with lower intra-class variability and higher identity separability tend to produce more compact genuine score distributions and more separated impostor distributions, resulting in higher similarity scores and consequently stricter thresholds. Conversely, datasets with increased variability or reduced separability exhibit broader and overlapping score distributions, leading to lower thresholds. 

This behavior is consistent with our observations: Syn-Multi-PIE produces systematically higher thresholds, indicating tighter clustering of identity representations, while FLUXSyn-ID yields lower thresholds, suggesting greater variability and a distribution closer to real-world conditions. These findings highlight that threshold calibration is not only dependent on dataset scale, but critically on how well synthetic data reproduces the balance between intra-class variability and inter-class separability observed in real-world data.

\subsection{Evaluation results}
We evaluate threshold transferability across datasets by applying calibration thresholds derived from synthetic data to real-world evaluation sets with different characteristics. Color FERET represents a controlled scenario with low intra-class variability and well-separated score distributions (Fig.~\ref{fig:real_distributions}a), while CFPW introduces unconstrained conditions with increased variability and overlap between genuine and impostor distributions (Fig.~\ref{fig:real_distributions}b). This setup enables us to investigate how differences in score distribution shape affect the stability of calibrated thresholds, particularly whether synthetic calibration preserves target FMR and acceptable FNMR under domain shift at increasingly stringent operating points.

\begin{table}[H]
\begin{center}
\setlength{\tabcolsep}{1.5pt}
\begin{tabular}{lccccc}
\hline
Calibration & 0.10\% & 0.05\% & 0.01\% & 0.005\% & 0.001\%  \\
\hline
\multicolumn{6}{l}{Observed FNMR under different calibrations.} \\
\hline
FLUXSyn. & 0.000 & 0.000 & 0.027 & 0.055 & 0.219 \\
Syn-Multi. & 0.329 & 0.657 & 1.972 & 2.957 & 5.942\\
Combined & 0.000 & 0.082 & 0.548 & 0.904 & 2.574\\
\hline
\multicolumn{6}{l}{Observed FMR under different calibrations.} \\
\hline
FLUXSyn.    & 0.0007\% & 0.0005\% & 0.0003\% & 0.0003\% & *\\
Syn-Multi. & 0.0002\% & *        & *        &*         & *\\
Combined      & 0.0003\% & 0.0003\% & 0.0002\% & *        & *\\
\hline
\end{tabular}
\end{center}
\caption{\label{tab:fnmr}FNMR (\%) and observed FMR (\%) (*$<$ 0.0002\%) on Color FERET under different calibration strategies.}
\end{table}

\noindent\textbf{Synthetic-to-real transfer.}
As shown in Table~\ref{tab:fnmr}, thresholds derived from FLUXSyn-ID generalize reasonably well to Color FERET, maintaining low FNMR across operating points while preserving very low observed FMR. In contrast, Syn-Multi-PIE calibration leads to substantially higher FNMR, indicating poor transferability. This suggests that not all synthetic datasets are equally suitable for threshold calibration.

\noindent\textbf{Unconstrained evaluation.}
Table~\ref{tab:fnmr_CFPW} shows that all calibration strategies experience significant degradation on CFPW, reflecting the increased variability and overlap in score distributions (Fig.~\ref{fig:real_distributions}b). While FLUXSyn-ID maintains relatively lower FNMR compared to other synthetic calibrations, performance deteriorates rapidly at stricter operating points, highlighting the sensitivity of threshold calibration under realistic conditions.

\begin{table}[H]
\begin{center}
\setlength{\tabcolsep}{2.2pt}
\begin{tabular}{lccccc}
\hline
Calibration & 0.10\% & 0.05\% & 0.01\% & 0.005\% & 0.001\%\\
\hline
\multicolumn{6}{l}{Observed FNMR under different calibrations.} \\
\hline
FLUXSyn.   & 6.69\%  & 8.69\%  & 14.48\% & 17.48\% & 25.79\%\\
Syn-Multi. & 28.47\% & 34.52\% & 48.78\% & 55.15\% & 67.62\%\\
Combined   & 14.21\% & 19.05\% & 32.77\% & 38.89\% & 53.01\%\\
\hline
\multicolumn{6}{l}{Observed FMR under different calibrations.} \\
\hline
FLUXSyn.   & 0.0005\%&0.0002\% & *       & *       &*          \\
Syn-Multi. & *      & *        & *       & *       &*          \\
Combined   & *      & *        & *       & *       &*          \\
\hline
\end{tabular}
\end{center}
\caption{\label{tab:fnmr_CFPW}FNMR (\%) and observed FMR (\%) (*$<$ 0.0002\%) on CFPW under different calibration strategies.}
\end{table}

Across Tables~\ref{tab:fnmr} and \ref{tab:fnmr_CFPW}, stricter operating points (e.g., 0.001\%) reveal strong sensitivity to small shifts in score distributions. Minor differences in impostor tails lead to disproportionately large increases in FNMR, demonstrating the difficulty of stable calibration in operational low-FMR regimes.

\noindent\textbf{Synthetic-to-synthetic transfer.}
Results in Tables~\ref{tab:fnmr_Syn} and \ref{tab:fnmr_Flux} show strong asymmetry and poor transferability between synthetic datasets. Thresholds derived from FLUXSyn-ID maintain relatively low FNMR when applied to Syn-Multi-PIE but produce significantly elevated FMR, indicating over-permissive thresholds. Conversely, Syn-Multi-PIE calibration leads to large FNMR increases when applied to FLUXSyn-ID, reflecting overly strict thresholds. These results confirm that synthetic datasets exhibit fundamentally different score distributions.

\begin{table}[H]
\begin{center}
\setlength{\tabcolsep}{3pt}
\begin{tabular}{lccccc}
\hline
Calibration & 0.10\%  & 0.05\%  & 0.01\% & 0.005\% & 0.001\%\\
\hline
\multicolumn{6}{l}{Observed FNMR.} \\
FLUXSyn.    & 0.073\% & 0.170\% & 0.787\%& 1.330\% & 3.687\%\\
\hline
\multicolumn{6}{l}{Observed FMR.} \\
FLUXSyn.   & 2.390\%  & 1.558\% & 0.574\%& 0.370\% & 0.132\%\\
\hline
\end{tabular}
\end{center}
\caption{\label{tab:fnmr_Syn}FNMR (\%) and observed FMR (\%) on Syn-Multi-PIE evaluation dataset using thresholds calibrated on FLUXSyn-ID dataset.}
\end{table}

\begin{table}[H]
\begin{center}
\setlength{\tabcolsep}{2.3pt}
\begin{tabular}{lccccc}
\hline
Calibration & 0.10\%  & 0.05\%  & 0.01\% & 0.005\% & 0.001\%\\
\hline
\multicolumn{6}{l}{Observed FNMR.} \\
Syn-Multi.  & 0.208\% & 0.535\% & 3.066\%& 5.36\% &13.96\%\\
\hline
\multicolumn{6}{l}{Observed FMR.} \\
Syn-Multi.  &0.0006\%  &0.0002\% & * & * &* \\
\hline
\end{tabular}
\end{center}
\caption{\label{tab:fnmr_Flux}FNMR (\%) and observed FMR (\%) (*$<$ 0.0002\%) on FLUXSyn-ID evaluation dataset using thresholds calibrated on Syn-Multi-PIE dataset.}
\end{table}

\subsection{Evaluation with morphed images}

Table~\ref{tab:morp} shows that calibration strategy significantly impacts vulnerability to morph attacks. Thresholds derived from FLUXSyn-ID achieve low FNMR but result in high NMPR and FMPR at relaxed operating points, indicating increased susceptibility. Conversely, stricter thresholds reduce morph acceptance but at the cost of usability. This highlights a fundamental trade-off between recognition performance and security.

\begin{table}[H]
\begin{center}
\setlength{\tabcolsep}{2.5pt}
\begin{tabular}{l|lc|lc|lc}
\hline
        & FLUX  &       & SYN   &       & Comb. &      \\     
FMR     & NMPR  & FMPR  & NMPR  & FMPR  & NMPR  & FMPR  \\
\hline
0.10\%  & 0.997 & 0.734 & 0.834 & 0.136 & 0.972 & 0.416 \\
0.05\%  & 0.994 & 0.634 & 0.748 & 0.076 & 0.937 & 0.299 \\
0.01\%  & 0.969 & 0.408 & 0.503 & 0.018 & 0.771 & 0.088 \\
0.005\% & 0.950 & 0.328 & 0.405 & 0.007 & 0.679 & 0.051 \\
0.001\% & 0.560 & 0.025 & 0.213 & 0.002 & 0.434 & 0.008 \\
\hline
\end{tabular}
\end{center}
\caption{\label{tab:morp}Morph passing ratio per calibration dataset and FMR. NMPR = Nearer match passing rate, FMPR = farther match passing rate}
\end{table}

\subsection{Summary of findings}
The results demonstrate three key observations. First, threshold calibration using synthetic data is highly dataset-dependent, with substantial differences in transfer performance across synthetic datasets. Second, thresholds do not generalize reliably even between synthetic datasets, indicating fundamental differences in score distributions. Third, these discrepancies are amplified at low-FMR operating points, where small distribution shifts result in large performance degradation. Together, these findings indicate limitations of synthetic data for operational threshold calibration, particularly under domain shift and low-FMR constraints.

\section{Discussion}
The results provide a systematic view of how threshold calibration using synthetic data behaves under domain shift, particularly at low-FMR operating points. In this section, we interpret these findings in the context of dataset characteristics, operational constraints, and security implications for real-world deployment.

\subsection{Limitations of Synthetic Data for Threshold Calibration}

The primary limitation of synthetic data for threshold calibration lies in mismatches in score distributions, particularly in impostor tail regions that determine low-FMR operating points. While synthetic datasets approximate central tendencies of genuine and impostor scores, calibration depends on extreme quantiles of the impostor distribution. As illustrated in Fig.~\ref{fig:distributionsFLUX} and Fig.~\ref{fig:real_distributions}, even small discrepancies in tail behavior lead to substantial shifts in estimated thresholds. This sensitivity is further amplified by statistical uncertainty in the estimation of extreme impostor quantiles, which is inherently challenging even with large datasets.

This effect is evident across synthetic datasets. Syn-Multi-PIE produces systematically higher thresholds due to tightly clustered identity representations, while FLUXSyn-ID yields lower thresholds that better align with real-world data. These findings suggest that datasets optimized for controlled variability or benchmarking may fail to capture the variability present in operational environments, leading to miscalibrated decision boundaries. This limitation is particularly critical in border control applications such as EES.

\subsection{Dataset Characteristics and Transferability}

As shown in Tables~\ref{tab:fnmr} and \ref{tab:fnmr_CFPW}, thresholds derived from FLUXSyn-ID generalize more effectively than those from Syn-Multi-PIE, yet still degrade under domain shift. In practice, operational EES conditions lie between controlled (FERET-like) and unconstrained (CFPW-like) extremes, combining standardized capture with residual variability.

The evaluation on CFPW demonstrates that increased variability and overlap between genuine and impostor distributions significantly affect calibration stability. In such conditions, thresholds estimated from synthetic data fail to preserve both target FMR and acceptable FNMR.

In contrast, results on Color FERET show that synthetic calibration can be effective when the target domain exhibits controlled acquisition conditions and well-separated score distributions. Thresholds derived from FLUXSyn-ID transfer with minimal degradation, maintaining low FNMR and low observed FMR. This suggests that when operational conditions align with synthetic data characteristics, calibration can be reasonably approximated. This is particularly relevant for EES, where capture is partially standardized under ICAO-compliant conditions, but still subject to operational variability.

\subsection{Implications for Operational Systems}

Calibration based solely on synthetic data may produce overly permissive or overly strict thresholds. Overly permissive thresholds increase false match risk, while overly strict thresholds degrade usability. Given stringent low-FMR requirements, even minor calibration errors can translate into significant operational risks. Our results indicate that synthetic data alone may be insufficient under domain shift or unconstrained conditions.

The morph attack evaluation reveals a trade-off between recognition performance and security. As shown in Table \ref{tab:morp}, calibration strategies that achieve low FNMR tend to produce higher morph acceptance at relaxed operating points, while stricter thresholds reduce vulnerability at the cost of rejecting genuine users. This highlights that calibration must consider adversarial risks, not only recognition accuracy. This further illustrates that threshold calibration errors induced by distribution mismatch directly translate into exploitable security vulnerabilities.

\subsection{Role of Synthetic Data in Calibration Pipelines}

Despite these limitations, synthetic data remains valuable for system development and controlled analysis. However, it should be used for preliminary calibration rather than final threshold selection. Reliable deployment requires validation using representative real-world data. Hybrid approaches combining synthetic and representative real-world data may provide a practical compromise. This recommendation is consistent with recent European policy guidance, which recognizes the value of synthetic biometric datasets while emphasizing the continued importance of representative operational data for biometric accuracy evaluation \cite{eulisa2026biometric}.

\noindent\textbf{Practical recommendations}

\begin{itemize}
    \item Use synthetic data for initial calibration and system development, but not for final threshold selection.
    \item Validate thresholds on in-domain real data whenever possible.
    \item Incorporate safety margins, especially at low-FMR operating points.
    \item Evaluate robustness under adversarial conditions such as morph attacks.
\end{itemize}

\section{Conclusion}
We investigated the use of synthetic face datasets for threshold calibration under operational constraints. Threshold calibration is highly sensitive to dataset characteristics, and thresholds calibrated on one dataset may not reliably generalize across domains, particularly under distribution shift. Even between synthetic datasets, differences in score distributions lead to inconsistent calibration outcomes.

Accurate modeling of impostor score tails is critical for reliable low-FMR operation. Synthetic datasets, while effective for training and benchmarking, do not consistently reproduce these properties under unconstrained or shifting conditions.

These limitations have direct implications for systems such as EES, where small calibration errors translate into security or usability risks. Calibration also impacts robustness to morph attacks, revealing a trade-off between performance and security.

Synthetic data remains useful for system development and preliminary calibration, but reliable deployment requires validation with real-world or high-quality synthetic data generated to match real-world data distributions appropriate to the target context. Future work will explore hybrid calibration strategies and improved distributional alignment.

An important direction for future work is the development of synthetic datasets that more faithfully capture operational variability in biometric acquisition. In particular, our results suggest that improving the modeling of acquisition conditions in border control, such as differences between self-service kiosks, guided capture systems, and mobile devices, may be critical for aligning score distributions, especially in the impostor tails relevant for low-FMR calibration. Synthetic data generation frameworks that incorporate such variability in a controllable and realistic manner could improve the reliability of calibration and evaluation, provided that their alignment with real-world distributions is carefully validated.

\section{Limitations}

This study has several limitations. First, experiments were conducted using a single model (EdgeFace), and results may vary with higher-capacity architectures. Second, evaluation is limited to specific datasets and may not capture full operational diversity. Third, while morph attacks were considered, other adversarial scenarios were not explored. Finally, deployment factors such as sensor variability and acquisition noise were not explicitly modeled.

Further validation across models, datasets, and operational conditions is required.

{\small
\bibliographystyle{ieee}
\bibliography{egbib}
}

\end{document}